%% file: neurips_2019-ccal.tex
\documentclass{article}

 \usepackage[preprint]{neurips_2019}

\usepackage[utf8]{inputenc} 
\usepackage[T1]{fontenc}    
\usepackage{hyperref}       
\usepackage{url}            
\usepackage{booktabs}       
\usepackage{amsfonts}       
\usepackage{nicefrac}       
\usepackage{microtype}      

\usepackage{graphicx}
\usepackage{amsmath}
\usepackage{mathrsfs}
\usepackage{amsthm}

\setcitestyle{numbers}

\newtheorem{lemma}{Lemma}
\newtheorem{proposition}{Proposition}

\def\R{{\bf R}}
\def\N{{\bf N}}
\def\F{\mathscr{F}}
\def\S{\mathscr{S}}
\def\A{\mathscr{A}}

\DeclareMathOperator{\Th}{Th}
\DeclareMathOperator{\rank}{rank}

\def\x{\xi}
\def\s{\zeta}
\def\W{M}
\def\w{m}
\def\outputs{\eta}
\def\tvar{\tau}

\def\|#1|{\vbox{\hbox{\includegraphics{./const-#1.mps}}}}
\def\<#1>{\hbox{\includegraphics{./const-#1.mps}}}

\title{Spatiotemporal Local Propagation}

\author{%
  Alessandro Betti\\
  University of Florence\\
  Florence, Italy\\
  \texttt{alessandro.betti@unifi.it}\\
  \And
  Marco Gori\\
  SAILab, University of Siena\\
  Siena, Italy\\
  \texttt{marco@diism.unisi.it}\\
}

\begin{document}

\maketitle

\begin{abstract}
This paper proposes an in-depth re-thinking of neural computation
that parallels apparently unrelated laws of physics, that are formulated
in the variational framework of the least action principle.
The theory holds for neural networks that are also based on any digraph, 
and the resulting computational scheme exhibits the intriguing property of 
being truly biologically plausible. The scheme, which is referred to as SpatioTemporal
 Local Propagation (STLP), is local in both space and time.
Space locality comes from the expression of the network connections
by an appropriate Lagrangian term, so as the corresponding
computational scheme does not need the backpropagation (BP) of the error,
while temporal locality is the outcome of the variational
formulation of the problem. Overall, in addition to conquering the often invoked 
biological plausibility missed by BP, 
the locality in both space  and time that arises from the proposed theory
can neither be exhibited by Backpropagation Through Time (BPTT) 
nor by Real-Time Recurrent Learning (RTRL).
\end{abstract}

\section{Introduction}
\input intro.tex

\section{Modified Dirichlet problem}
\label{dirichlet-sec}
\input dirichlet.tex


\section{Neural network constraints}
\label{constraint-sec}
\input constraints.tex

\section{Cognitive action and laws of learning: Feedforward architecture}
\label{law-sec}
\input laws.tex

\section{Conclusions}
\input conclusions.tex

\bibliography{nips19,nn}
\bibliographystyle{unsrt}

\end{document}

%% file: intro.tex
Since mid-eighties, the explosion of interest in neural computation 
has been mostly fueled by finite-dimensional optimization in the space
of the connection weights. Because of the typical large number of parameters
involved, gradient-descent methods have dominated the searching 
heuristics. Moreover, it early became clear that  large-scale problems 
can only be faced thanks to stochastic gradient descent (SGD) and related algorithms, 
that represent the on-line side of classic batch-mode optimization schemes.
To some extent, SGD gives learning a sort of temporal dimension.
If one assumes that the examples come at discrete time then
weight updating takes place, for each example, at each temporal step.
As early pointed out in the seminal PDP book~\cite{Rumelhart86a} (p.324), on-line learning can be regarded as an approximation of gradient descent
of the error function. In their words
\begin{quote}
\small {\em
	By changing the weights after each pattern is presented we 
	depart to some extent from a true gradient descent in $E$.
	Nevertheless, provided the learning rate (i.e. the constant of 
	proportionality) is sufficiently small, this departure will be negligible 
	and the delta rule will implement a ver close approximation to gradient-descent
	in sum-squared error.}
\end{quote} 
The resulting on-line process, along with mini-batch versions, have been
the subject of  in-depth recent investigations 
(e.g.~\cite{BottouCN18}) that 
have contributed to shed light on SGD and on many specific versions that
have been massively using in machine learning. 

This paper is motivated by the curiosity of providing a truly new foundation of
learning, where  ``time'' is regarded as an intrinsic variable for the acquisition of
concepts.  Basically, we propose a formulation of learning by differential equations instead of by the dominating approach of using
finite-dimensional optimization. This can be traced back to 
a number of relevant contributions, including~\cite{Pineda87}
and~\cite{Pearlmutter89b}, as well as to a recent interesting
continuous-based formulation of deep learning~\cite{conf/nips/ChenRBD18}.
While following this track, this paper
proposes a new view of learning, that can regarded as the outcome
of laws of nature. 
We use the unifying view that arises from physics when
using variational calculus and, particularly, when deriving laws of nature as
stationary points of the action. We establish a parallel with  mechanics
according to which particle positions is associated with the neural parameters
(weights and outputs), so as the velocity turns out to indicate the rate of
the learning process (see Table~\ref{table}). The kinetic energy has related meaning,
while the potential energy indicates the degree of satisfaction of the environmental
constraints -- for instance, in case of supervised learning the potential turns out to
be a loss function. 
Interestingly, the presence of motion constraints on particles has a counterpart
in the constraints that express the neural model, so as ``learning motion'' is
a stationary point of a functional, referred to as the {\em cognitive action}, 
under the neural architectural constraints. Like in mechanics, the resulting solution
is a differential equations in the Lagrangian variables that dictates the evolution of 
the weights and of the neural outputs. The learning behavior reminds us of
damped oscillators and the process of dissipation leads to ordered configurations which
correspond to the outcome of learning. As dissipation increases the proposed  theory
leads to solutions that approaches classic gradient descent.

The most striking results of the theory are deeply rooted 
into the variational formulation under the subsidiary conditions, that represent the neural constraints. 
It is shown that STLP exhibits locality in both space and
time for neural networks defined by any digraph. Temporal locality is basically the
outcome of the variational optimization that yields models based on differential equation.
Interestingly,  space locality turns out to be the outcome of imposing the stationarity of the cognitive action under neural constraints. 
The issue of biological plausibility has been recently
the subject of a related investigation in~\cite{DBLP:journals/corr/BengioLBL15}. 
 
The message that emerges from the paper is that in order to
gain a truly biological plausibility,  temporal locality and
strong space locality must be supported. On the other hand,
classic algorithms for gradient computation in recurrent neural 
network do not exhibit this property:
neither BPTT (BackPropation
Through Time) nor Real-Time Recurrent Learning (RTRL) possess space and temporal
locality. BPTT is local in space, but not in time, whereas RTRL is local in time, but not 
in space~\cite{Williams89b}. 
Moreover, in these classic algorithms, space locality refers to the property gained by the backpropation
factorization, not to strong space locality that is gained by STLP. 
Overall,  the proposed theory stimulates a re-thinking of neural 
computation driven by laws of nature, where there is no distinction
between learning and test, where the weight updating is paired
with computation of the output in the learning environment.
The theory also opens the doors for an in-depth reformulation of learning algorithms. 

%
%


%% file: dirichlet.tex
Let $\Omega$ be an open, bounded domain in $\R^n$, let
$u\colon\Omega\to \R^N$, $\varpi\in{\cal
C}^1(\Omega; (0,+\infty))$ and $\F(u):= \int_\Omega F(x,u,\nabla u)\,dx$ be, and
define the following functional
\begin{equation}
\S(u):=\frac{1}{2}\int_\Omega \vert\nabla u(x)\vert^2\, \varpi(x)dx+\F(u),
\label{general-functional}\end{equation}
which is a weighted Dirichlet integral plus the $\F(u)$ term. We are here
interested in the necessary conditions for $u$ to be an extremizer of the
{\it modified Dirichlet functional} \eqref{general-functional} subject to
a class of  holonomic constraints of the form
\footnote{In this section, for the sake of simplicity, we are considering
holonomic constraints that do not depend explicitly on the independent
variable $x$, however the arguments presented here can be readily generalized
also to include the case $G(x,u(x))=0$. Indeed in
Section~\ref{constraint-sec} we will consider general holonomic constraints.
} $G(u(x))=0$. Let us consider the problem in Eq.~\eqref{general-functional} where
$u$ is subject to the constraints $G(u(x))=0$ for all $x\in \overline\Omega$
and $G(z)$ of class ${\cal C}^2(\R^N, \R^r)$. Furthermore assume that
the $r\times N$ Jacobian  matrix defined by  $G_z=(G^i_{z^j})$
satisfies\footnote{We could ask for a less restrictive condition here, namely that
$G_z(z)$ should be full rank on all the points $z\in\R^N$ such that
$G(z)=0$.}:
\begin{equation}
\rank G_z=r\quad\hbox{for
all  $z\in \R^N$}.
\label{full-rank-condition}
\end{equation}
From the theory of calculus of variation with subsidiary conditions
(see \cite{giaquita-hildebrandt} Chap.~2) we know that there exist
$\lambda_1,\dots,\lambda_r\in {\cal C}^0(\Omega)$ such that the
constrained stationary points of $\S$ coincides with the unconstrained
stationary points of the extended functional
\begin{equation}
\S^*(u)={1\over 2}\int_\Omega \vert \nabla u(x)\vert^2\varpi(x)-
\lambda_j(x)G^j(u(x))\, dx+\F(u),
\label{const-func-gen}\end{equation}
and the Euler equation for this functional are
\footnote{throughout this paper we will adopt Einstein summation convention;
that is to say two repeated indices,
unless otherwise specified imply summation.}
\begin{equation}
-\varpi\Delta u -u_{x^\alpha}\varpi_{x^\alpha}-\lambda_\ell G^\ell_z(u)
+L_F(u)=0,\label{ELE-gen-mult}\end{equation}
where $L_F(u)$ is the Euler operator (\cite{giaquita-hildebrandt} p. 18 and
\cite{courant-hilbert}).
Differentiating the constraints two times with respect to
$x^\alpha$ one obtaines
\begin{equation}
-\Delta u\cdot G_z^j(u)=G^j_{z^iz^k}(u)u^i_{x^\alpha}u^k_{x^\alpha},
\qquad 1\le j\le r.
\end{equation}
Hence if we scalar  multiply Euler equation by $G_z^j(u)$ we obtain
\begin{equation}A_{ij}(u)\lambda_\ell=\varpi
G^j_{z^iz^k}(u)u^i_{x^\alpha}u^k_{x^\alpha}-\varpi_{x^\alpha}(u_{x^\alpha}\cdot
G^j_z(u)) +L_F(u)\cdot G_z^j(u),
\label{mult-lin-eq}
\end{equation}
where we defined  $A_{j\ell}(u):=G_z^j(u)\cdot G^\ell_z(u)$.
Therefore  Euler equations for the constrained
functional~\eqref{const-func-gen}  are~\eqref{ELE-gen-mult}
with
\begin{equation}
\lambda_\ell=(A^{-1}(u))_{\ell j}\bigl(\varpi
G^j_{z^iz^k}(u)u^i_{x^\alpha}u^k_{x^\alpha}
-\varpi_{x^\alpha}(u_{x^\alpha}\cdot G^j_z(u))
+L_F(u)G_z^j(u)\bigr).
\label{mult-eq-gen}
\end{equation}
Notice that in order to get from Eq.~\eqref{mult-lin-eq} to
\eqref{mult-eq-gen} we need to know that
$A$ is invertible. Whenever our assumption~\eqref{full-rank-condition} holds
the Gram matrix $A(u)$ turns out to be a invertible in view of the following
(well known) lemma:

\begin{lemma}
If $v_1$,\dots, $v_n$ are $n$ linear independent vectors, then the Gram matrix
$G_{ij}:=(v_i,v_j)$ is positive definite.
\end{lemma}

\begin{proof}
$(x,Gx)=x_i(v_i,v_j)x_j=(x_i v_i, x_j v_j)=\Vert v_i x_i\Vert^2\ge 0$.
However $\Vert v_i x_i\Vert=0$ if and only if $x_iv_i=0$, therefore
we can conclude that $(x,Gx)>0$ for every $x\ne 0$.
\end{proof}

\begin{table}
\small
\begin{tabular}{ccl}  
\toprule
{\bf Learning}  & {\bf Mechanics} & {\bf Remarks} \\
\midrule
$(W,x)$       & $u$  & 
                       Weights and neuronal outputs
                       are interpreted as generalized
                       coordinates.\\
\noalign{\smallskip}
$(\dot W,\dot x)$& $\dot u$ & 
                              Weight variations and neuronal variations
                              are interpreted as generalized velocities.
                              \\
\noalign{\medskip}
  $\A(x,W)$ & $\S(u)$&  
                       The cognitive action is the dual of the action in
                       mechanics.
                       \\
\bottomrule
\end{tabular}
\medskip
\caption{Links between learning theory and classical mechanics.}
\label{table}
\end{table}


%% file: constraints.tex
The typical learning paradigm within the framework of NN consists of a
model, that depends on a set of parameters $W$, together with 
an update rule for the parameters; this rule is usually  a gradient
descent of a function that measure the goodness
of the model on a specific learning task. However in this section
we will show that when the dynamics of the parameters $W$ is described
by laws that comes from stationarity conditions of a functional, as
it happens for canonical coordinates in classical mechanics (see
Table~\ref{table}), then
the NN model can be treated using the theory of constraints described in the
previous sections. 
As an immediate  consequence of this approach
the learning process gains temporal and spatial locality even in the
case of recurrent NN.

\begin{figure}
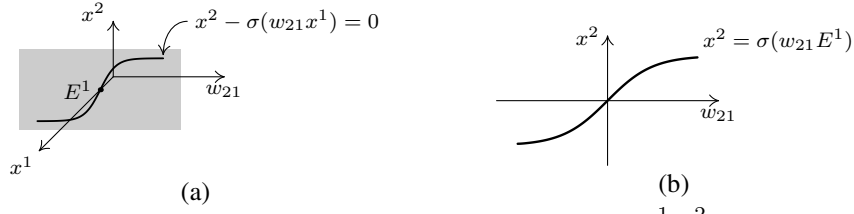

\[{\vcenter{\|5|}\atop\hbox{(a)}}\qquad\qquad {\vcenter{\|3|}\atop\hbox{(b)}}\]
\vskip -1pc
\caption{Visualization of the neural constraints
for the neural network \protect\<4> (one input $\{1\}$ 
and two neurons $\{1,2\}$).  Constraint $G^2(x,W)=0$, 
restricted to the plane $x^1=E^1$, is shown in (a). In (b), 
such restriction is represented in the  $w_{21}$--$x^2$ plane.}
\label{neuron-const-fig}
\end{figure}

First of all let us describe the architecture of the models that we will
address. Given a simple digraph $D=(V,A)$ of order $\nu$ without loss of
generality we can assume $V=\{1,2,\dots,\nu\}$ and $A=\{
(i,j)\in\N^2\mid i\in V, j\in V\}$. A neural network constructed on $D$
consists of a set of maps\footnote
{Please notice that now $x$ is a the variable of the variational
problem, and therefore represent a mapping $t\mapsto x(t)$. It not to be
intended as the independent variable of the problem described in the previous sections.} $i\in V\mapsto x^i\in \R$ and $(i,j)\in A\mapsto
w_{ij}\in \R$ together with $\nu$ constraints $G^j(x,W)=0$
$j=1,2,\dots \nu$ where $(W)_{ij}=w_{ij}$. Let ${\cal M}_\nu(\R)$ be the
set of all $\nu\times\nu$ real matrices and 
${\cal M}^{\downarrow}_\nu(\R)$ the set of all $\nu\times\nu$
strictly lower triangular matrices over $\R$. If
$W\in{\cal M}^{\downarrow}_\nu(\R)$ we say that the NN has a feedforward
structure. In this
paper we will consider both feedforward NN and NN with cycles. The relations
$G^j=0$ for $j=1,\dots, \nu$
 specify the computational scheme with which the information
diffuses trough the network. 
In a typical network with $\omega$ inputs these
constraints are defined as follows (see also Fig.~\ref{neuron-const-fig}): 
For any vector $\x\in\R^\nu$, for any matrix $\W\in{\cal
M}_\nu(\R)$ with entries $\w_{ij}$ and for any given ${\cal C}^1$ map 
$e\colon (0,+\infty)\to\R^\omega$ we define the constraint on neuron $j$
when the example $e(\tau)$ is presented to the network as
\begin{equation}G^j(\tau,\x,\W):=\begin{cases}
\x^j-e^j(\tau), & \text{if $1\le j\le \omega$};\\
\x^j-\sigma(\w_{jk} \x^k) & \text{if $\omega <j\le \nu$},
\end{cases}\label{neuron-constraints-structure}\end{equation}
where $\sigma\colon \R\to\R$ is of class ${\cal C}^2(\R)$.

Our goal here is to show that such relations, that normally are considered
just a local description of the compositional structure of the NN, once
properly interpreted as constraints in the space $x-W$
(see Fig.~\ref{neuron-const-fig})
are suitable
holonomic subsidiary conditions in the sense of~\eqref{full-rank-condition}.

Like in the case of classical mechanics, when dealing with learning processes
we are interested in the temporal dynamics of the variables when they are 
exposed to the data from which the learning is supposed to happen. For this
reason in this section we can restrict ourselves to the case $n=1$ and
regard this variable as time  ($x^1=t$). 
Moreover because the neural constraints $G^j(x,W)=0$ involve not only $W$
but also $x$ the $N$ variables $u_1,\dots, u_N$ split into
$x\in \R^\nu$ and $W\in{\cal M}_\nu(\R)$.

\medskip\noindent
{\bf Feedforward Networks.} Now let us consider the case
$W\in{\cal M}^{\downarrow}_\nu(\R)$
and let us  extend the theory described in Eq.~\eqref{general-functional}
by allowing
$\F(x,W):=\int F(t,x,\dot x,\ddot x,W,\dot W,\ddot W)\, dt$,
so that, in the end, we consider the functional
\begin{equation}
\A(x,W):=\int {1\over 2}(m_x\vert \dot x(t)\vert^2+m_W\vert \dot W(t)\vert^2)\,\varpi(t) dt+
\F(x,W),\label{cognitive-action}\end{equation}
subject to the constraints
\begin{equation}G^j(t,x(t),W(t))=0,\qquad 1\le j\le \nu.
\label{neuron-constraints}\end{equation}
Then the following proposition holds true:
\begin{proposition}
The matrix $({G_\x\atop \overline {G_\W}})\in {\cal M}_{(\nu^2+\nu)\times
\nu}(\R)$ is full rank.
\label{proposition}
\end{proposition}

\begin{proof}
First of all notice that if $(G_\x)_{ij}=G^j_{\x^i}$ is full rank
also $({G_\x\atop \overline {G_\W}})$ has this property. Then, since
\[G^j_{\x^i}(\tau,\x,\W)=\begin{cases}
\delta_{ij}, & \text{if $1\le j\le \omega$};\\
\delta_{ij}-\sigma'(\w_{jk} \x^k)\w_{ji} & \text{if $\omega <j\le \nu$,}
\end{cases}\]
we immediately notice that $G^i_{\x^i}=1$ and that for all $i>j$ we have
$G^i_{\x^i}=0$. This means that
\[
(G^j_{\x^i}(\tau,\x,\W))=\begin{pmatrix}
                          1&*&\cdots&*\\
                          0&1&\cdots&*\\
                          \vdots&\vdots&\ddots&\vdots\\
                          0&0&\cdots&1\end{pmatrix},\]
which is clearly full rank.
\end{proof}
Notice that this result heavily depends on the assumption
$W\in{\cal M}^{\downarrow}_\nu(\R)$; however we will now discuss how
the introduction of an additional variable that models the
degree of satisfaction of the neural constraints acts as a regularizer
of constraints~\eqref{neuron-constraints-structure} and ensure the satisfaction
of~\eqref{full-rank-condition}.

\medskip
\noindent
{\bf Recurrent networks.}
Let us  suppose that
we also assign to each neuron a variable $s$ that measure the degree of
violation of the constraint. Then Eq.~\eqref{neuron-constraints} assumes
the form $G^j(t,x,W,s)=0$, $j=1,2,\dots,\nu$ where
\begin{equation}
G^j(\tau,\x,\W,\s):=\begin{cases}
\x^j-e^j(\tau)+\s^j, & \text{if $1\le j\le \omega$;}\\
\x^j-\sigma(\w_{jk} \x^k)+\s^j & \text{if $\omega <j\le \nu$}.
\end{cases}\label{neuron-constraints-slack}\end{equation}
In doing so it is important to notice that Proposition~\ref{proposition} holds
without the assumption that $\W\in{\cal M}^\downarrow_\nu(\R)$ as it is
immediate to prove since $G^j_{\s^i}=\delta_{ij}$,
which is of course full rank. This important remark opens the possibility to
extend the theory to  networks with ``feedback'' connections
based on general simple digraphs.

In this formulation of the theory the action, described in Eq.~\eqref{cognitive-action} must be modified to take into account of the introduction of
the new variable $s$:
\begin{equation}
\A(x,W,s):=\int {1\over 2}(m_x\vert \dot x(t)\vert^2+m_W\vert \dot W(t)\vert^2
+m_s\vert \dot s(t)\vert^2)\,\varpi(t) dt
+\F(x,W,s),\label{cognitive-action-slack}
\end{equation}
where $\F(x,W,s):=\int F(t,x,\dot x,\ddot x,W,\dot W,\ddot W,s)\, dt$.


%% file: laws.tex
In the previous section we concentrated ourselves on showing that the set of
constraints that define a NN are good constraints (in the sense
of~\eqref{full-rank-condition}). In this section we  will focus on the
feedforward case described by the functional~\eqref{cognitive-action}
together with constraints~\eqref{neuron-constraints}. In particular we
will discuss the updates rules (Euler-Lagrange equations) for the variables $x$ and
$W$ derived from the stationarity conditions of the
functional~\eqref{cognitive-action}.
We notice in passing that when imposing the stationarity of
action $\delta \A=0$ we give rise to a computational model
that, in general, is remarkably different from classic 
optimization approaches used in machine learning, that
are typically driven by the gradient heuristics. Basically, 
the models arising from $\delta \A=0$, instead of gradually
reducing the action from its initial value,  satisfy this condition
for any time instant, thus resembling what happens for
Newtonian's laws.


We begin by deriving
the constrained Euler-Lagrange (EL) equations associated with the
functional~\eqref{cognitive-action} under
subsidiary conditions~\eqref{neuron-constraints}. The
constrained functional is
\begin{equation}
\A\ ^*(X,W)=\int {1\over 2}(m_x\vert \dot x(t)\vert^2+m_W\vert \dot W(t)\vert^2)\varpi(t)
-\lambda_j(t)G^j(t,x(t),W(t))\, dt+\F(x,W),
\end{equation}
and its EL-equations thus read

\begin{align}
  &-m_x\varpi(t)\ddot x(t)-m_x\dot\varpi(t) \dot x(t)
  -\lambda_j(t)G^j_\x(x(t),W(t))
+L^x_F(x(t),W(t))=0;\label{x-eq}\\
  &-m_W\varpi(t)\ddot W(t)-m_W\dot\varpi(t) \dot W(t)
  -\lambda_j(t)G^j_\W(x(t),W(t))
+L^W_F(x(t),W(t))=0,\label{W-eq}
\end{align}
where $L_F^x=F_x-d(F_{\dot x})/dt+d^2(F_{\ddot x})/dt^2$,
$L_F^W=F_W-d(F_{\dot W})/dt+d^2(F_{\ddot W})/dt^2$
are the functional derivatives of $F$ with respect to $x$ and $W$
respectively (see \cite{courant-hilbert}). An expression for Lagrange
multiplies, as it is explained in Section~\ref{dirichlet-sec}
is derived by differentiating two times the constraint with respect to the
time and using the obtained expression to substitute the second order terms
in the Euler equations. In this case the analogue of Eq.~\eqref{mult-lin-eq}
is
\begin{equation}
\begin{aligned}
\Bigl({G^i_{\xi^a}G^j_{\xi^a}\over m_x}+{G^i_{m_{ab}}G^j_{m_{ab}}
\over m_W}\Bigr)\lambda_j=&
\varpi\bigl(G^i_{\tvar\tvar}+2(G^i_{\tvar \x^a}\dot x^a
+G^i_{\tvar \w_{ab}}\dot w_{ab}
+G^i_{\x^a\w_{bc}}\dot x^a\dot w_{bc})\\
&+G^i_{\x^a\x^b}\dot x^a \dot x^b
+G^i_{\w_{ab}\w_{cd}}\dot w_{ab}\dot w_{cd}\bigr)\\
&-\dot \varpi( \dot x^a G^i_{\xi^a}+\dot w_{ab} G^i_{m_{ab}})
+{L^{x^a}_FG^i_{\xi^a}\over m_x}+{L^{w_{ab}}_FG^i_{m_{ab}}\over m_W},
\end{aligned}\label{mult-lin-eq-x-w}
\end{equation}
where $G^i_\tvar$, $G^i_{\tvar\tvar}$, $G^i_{\xi^a}$, $G^i_{\xi^a\xi^b}$,
$G^i_{\w_{ab}}$ and  $G^i_{\w_{ab}\w_{cd}}$ are the gradients and the
hessians of constraint~\eqref{neuron-constraints}.

\medskip
\noindent
{\bf Initial conditions. }
Suppose now that we want to solve Eq.~ \eqref{x-eq}--\eqref{W-eq}
with Cauchy initial conditions.
Of course we must choose $W(0)$ and $x(0)$ such that $g_i(0)\equiv 0$, where
we posed $g_i(t):= G^i(t, x(t), W(t))$, for $i=1,\dots,\nu$. However since the
constraint must hold also for all $t\ge0$ we must also have at least
$g'_i(0)=0$. These conditions written explicitly means
\[G^i_\tau(0,x(0),W(0))+G^i_{\xi^a}(0,x(0),W(0))\dot x^a(0)
+G^i_{m_{ab}}(0,x(0),W(0))\dot w_{ab}(0)=0.\]
If the constraints does not depend explicitly on time it is sufficient to
to choose $\dot x(0)=0$ and $\dot W(0)=0$, while for time dependent constraint
this condition leaves 
\[G^i_\tau(0,x(0),W(0))=0,\]
which is an additional constraint on the initial conditions $x(0)$ and $W(0)$
to be satisfied.
Therefore one possible consistent way to impose Cauchy conditions is
\begin{equation}
\begin{aligned}
&G^i(0,x(0),W(0))=0,\quad i=1,\dots,\nu;\\
&G^i_\tau(0,x(0),W(0))=0,\quad i=1,\dots,\nu;\\
&\dot x(0)=0;\\
&\dot W(0)=0.
\end{aligned}
\label{initial-cond}
\end{equation}
Higher derivative of $g_i(0)$ becomes automatically satisfied thanks
through the differential equations.


\medskip
\noindent
{\bf Supervised Learning and reduction to BP.}
In order to see how this theory can be readily
applied to learning let us restrict ourselves to the case
$W\in {\cal M}^\downarrow_\nu(\R)$ and choose $\varpi(t)=\exp(\vartheta t)$,
$\vartheta>0$, $m>0$.
Now let us choose
\[F(t,x(t),\dot x(t), \ddot x(t),W(t),\dot W(t),\ddot W(t))=-e^{\vartheta t}
V(x(t),y(t)),\]
where $y(t)$ is an assigned supervision signal and
\[V(x(t),y(t)):=\frac{1}{2}\sum_{i=1}^\outputs
(y^i(t)-x^{\nu-\outputs+i}(t))^2,\]
$x^{\nu-\outputs},\dots, x^{\nu}$ being the variables associated with
the outputs neurons. A typical input signal and the corresponding
supervision signal can be constructed from
a standard training set $\mathscr{L}:=\{(e_\kappa, d_\kappa)
\mid e_\kappa \in \R^\omega, d_\kappa\in\R^\eta, \kappa=1,\dots,\ell \}$
in the following manner. Choose a sequence of times $\langle t_n\rangle:=t_0,t_1,t_2,\dots$
such that $|t_{i+1}-t_i|=:\tau$ is constant $i\in\N$. Furthermore
define the following sequences: $\langle E_n\rangle:=
e_1,\dots, e_\ell,e_1,\dots e_\ell,\dots$ and
$\langle y_n\rangle:= d_1,\dots, d_\ell,d_1,\dots d_\ell,\dots$.
Let $R(t):= \sum_{n=0}^\infty \rho_\epsilon(t-t_n)$, where
$\rho_\epsilon(\cdot)$ are standard Friedrichs mollifiers and define
\[\bar E(t):=\sum_{n=0}^\infty E_n \chi_{[t_{n-1},t_n]}(t),\quad
\bar y(t):=\sum_{n=0}^\infty y_n \chi_{[t_{n-1},t_n]}(t),\]
where $\chi_A$ is the characteristic function of the set $A$ and
$t_{-1}=0$.
Then the signal
\[E(t):=(\bar E *R)(t),\quad{\rm and}\quad y(t):=(\bar y *R)(t),\]
is piecewise constant signals with smooth transitions. The temporal
behaviour of these signals is depicted in the side figure.
%

\smash{\raise-2pc\rlap{\kern23pc\vbox{
\hbox{\includegraphics{./qpe-3.mps}}}}}
\parshape 3 0pt 22pc 0pt 22pc 0pt \hsize
To understand the behaviour of the Euler equations~\eqref{x-eq}
and~\eqref{W-eq} 
we observe that in the case of feedforward networks, as it
is well known, the constraints
$G^j(t,x,W)=0$ can be solved for $x$ so that eventually we can express the
value of the output neurons in terms of the value of the input neurons.
If we let $f^i_W(e(t))$ be the value of $x^{\nu-i}$ when $x^1=e^1(t),\dots
,x^\omega=e^\omega(t)$, then the theory defined by \eqref{cognitive-action} under
subsidiary conditions~\eqref{neuron-constraints} is equivalent, when
$m_x=0$, to the 
unconstrained theory defined by
\begin{equation}
\int e^{\vartheta t}\Bigl(\frac{m_W}{2}\vert\dot W\vert^2-\overline V(t,W(t))
\Bigr)\, dt\label{supervised-unconstrained}
\end{equation}
where $\overline V(t,W(t)):=\frac{1}{2}\sum_{i=1}^\outputs
(y^i(t)-f^i_W(E(t)))^2$. The Euler equations associated
with~\eqref{supervised-unconstrained} are
\begin{equation}
\ddot W(t)+\vartheta \dot W(t)=-\frac{1}{m_W}\overline V_W(t,W(t)),
\label{euler-backprop}
\end{equation}
that in the limit $\vartheta\to\infty$ and $\vartheta m\to \gamma$ reduces to
the gradient method
\begin{equation}
\dot W(t)=-\frac1\gamma \overline V_W(t,W(t)),
\label{gradient-like-w}
\end{equation}
with learning rate $1/\gamma$.
Notice that the presence of the term $\varpi(t)$ that we proposed in the
general theory it is essential in order
to have a learning behaviour as it produce dissipation.

Typically the term $\overline V_W(t,W(t))$ in Eq.~\eqref{gradient-like-w}
can be evaluated using the Backpropagation algorithm; we will now show that
Eq.~\eqref{x-eq}--\eqref{mult-lin-eq-x-w} in the same limit used above
$m_x\to 0$, $m_W\to 0$, $m_x/m_W\to 0$ reproduces Eq.~\eqref{gradient-like-w}
where the term $\overline V_W(t,W(t))$ explicitly assumes the form
prescribed by BP. In order to see this choose $\vartheta=\gamma/m_W$ and
multiply both sides of Eq.~\eqref{x-eq}--\eqref{mult-lin-eq-x-w} by
$\exp(-\vartheta t)$, then take the limit
$m_x\to 0$, $m_W\to 0$, $m_x/m_W\to 0$. In this limit Eq.~\eqref{W-eq} and
Eq.~\eqref{mult-lin-eq-x-w} becomes respectively
\begin{align}
&\dot W=-\frac1\gamma  \sigma'(w_{ik}x^k) \delta_i x^j;\label{BP-grad}\\
&G^i_{\xi^a} G^j_{\xi^a} \delta_j=-V_{x^a}G^i_{\xi^a},\label{backward}
\end{align}
where $\delta_j$ is the limit of $\exp(-\vartheta t)\lambda_j$.
Because the matrix $G^i_{\x^a}G^j_{\x^a}$ not only is invertible,
but it is a Gram matrix if we define $T_{ij}:= G^j_{\xi^i}$, then
we have $G^i_{\x^a}G^j_{\x^a}=(T'T)_{ij}$. If we then  pose
$v_i:=-\gamma G^i_{\x^a}\dot x^a$, the $\delta$'s  satisfies
$T'T\delta=v$ with $T$ that is an upper triangular matrix. Solving this
equation is equivalent to the solution of $T'\mu= v$ and $T\delta=\mu$.
From this last equation we immediately see that once $\mu$ is known
$\delta$ is recursively derived starting from the output neurons.
Finally, we can interpret $\delta_i$
in  Eq.~\eqref{backward} as the delta-error, which is the 
recursively determined by Eq.~\eqref{BP-grad} because
of the special structure of of matrix $G^i_{\xi^a} G^j_{\xi^a}$.



\medskip
\noindent
{\bf Optimal inversion of $A=G^i_{\xi^a} G^j_{\xi^a}$}. 
Since $A$ is Gram matrix, its inversion of $A$, which is required for determining $\delta_{j}$ in Eq.~\ref{mult-lin-eq-x-w}, can be
efficiently determined with an optimal 
complexity~\cite{FraDah11}. Basically, we only need 
a number of dominant floating-point operation with grows
quadratically with the dimension of $A$.

\subsection{Simulation of the dynamics}
\begin{figure}[t]
\hbox to\hsize{
\includegraphics[width=0.5\hsize]{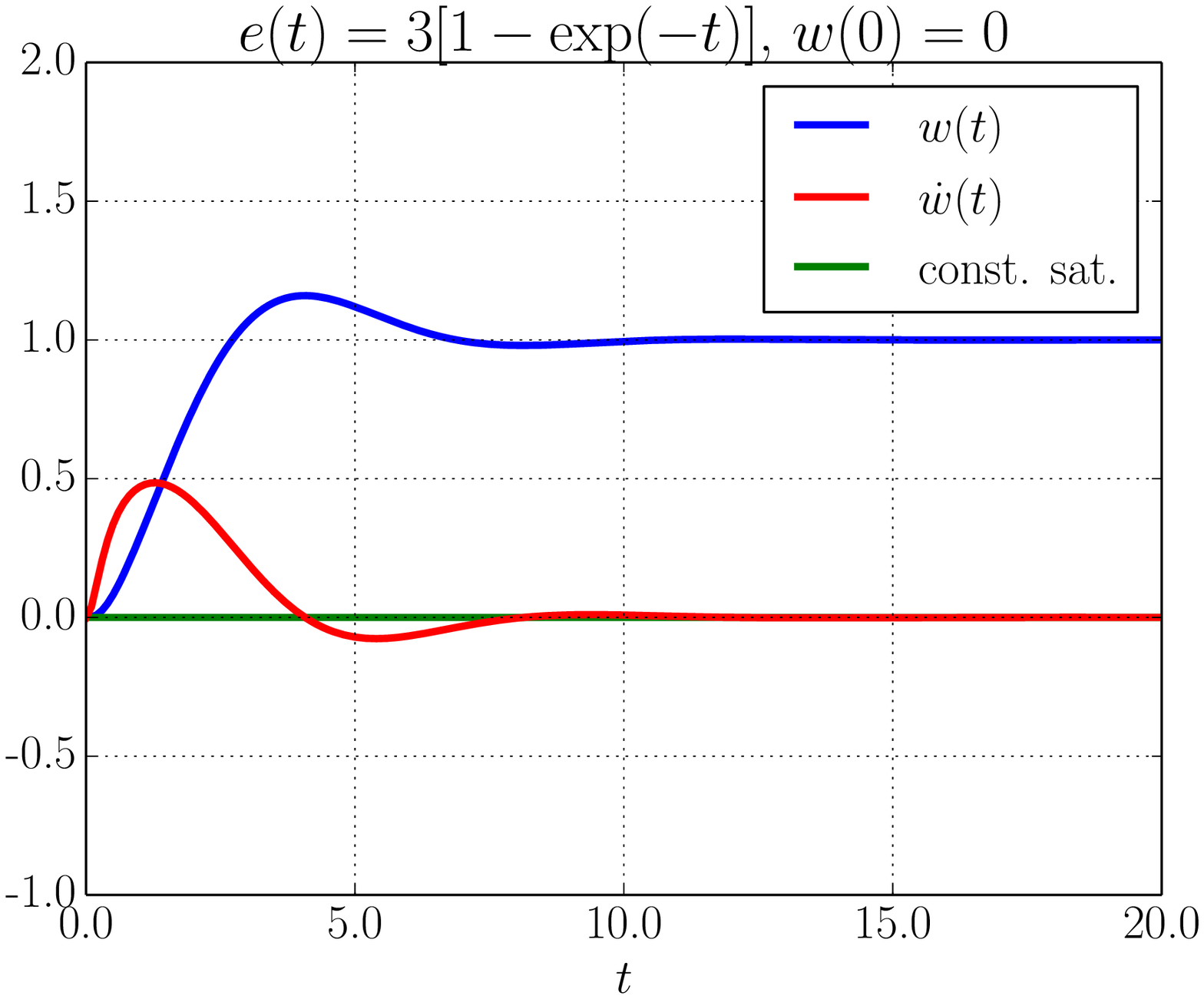}\hfill
\includegraphics[width=0.5\hsize]{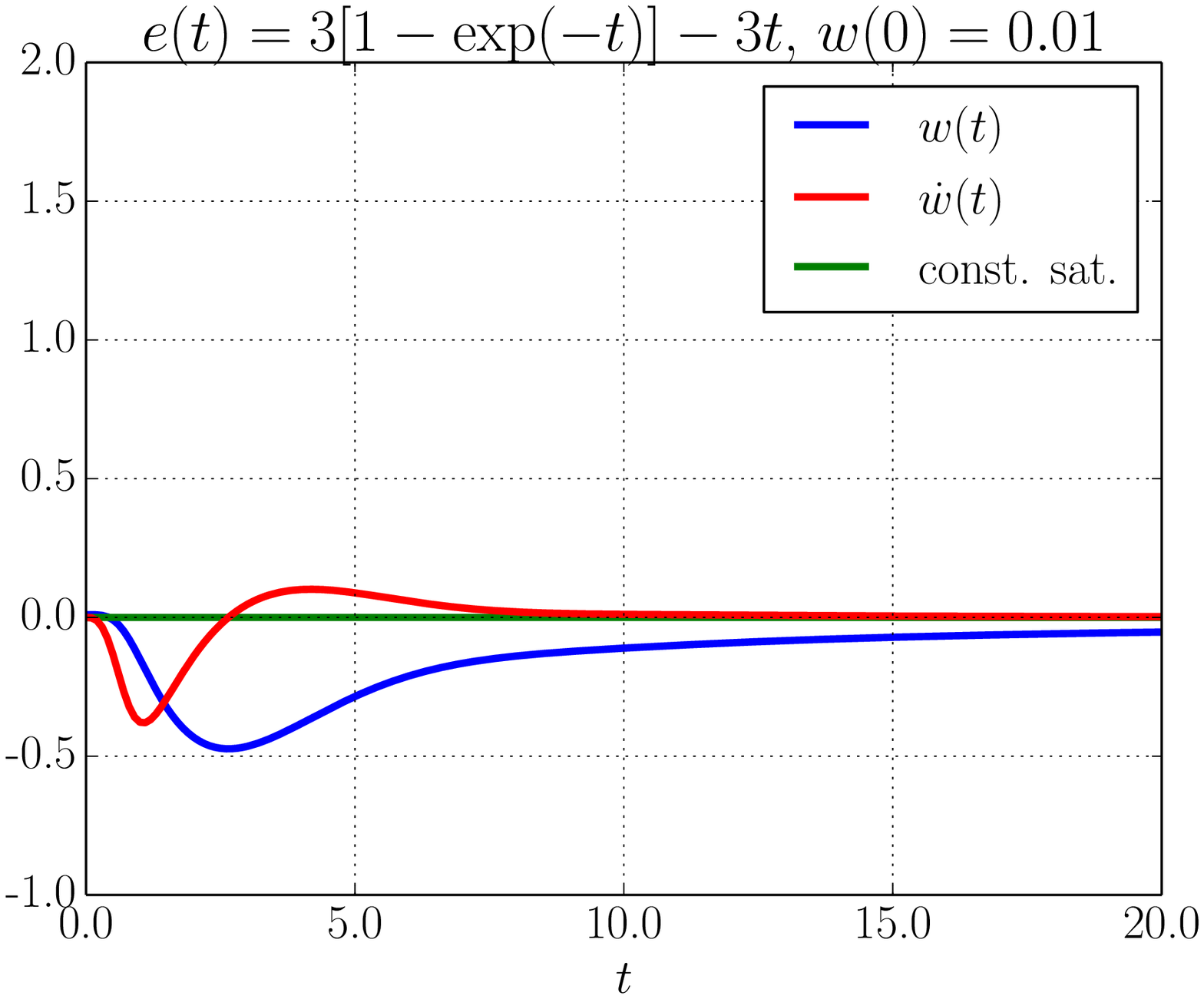}
}
\hbox to \hsize{\hfil(a)\qquad\quad\hfil\quad\qquad(b)\hfil}
\caption{Temporal evolution for the weights of a linear neuron
characterized by the constraint $G=x(t)-w(t)e(t)$. We used
$\overline V=(x-3)^2/2$ and we fixed the parameters as follows
$m_x=m_W=\vartheta=1$. The shown trajectories correspond to
the initial conditions $\dot w(0)=\dot x(0)=x(0)=0$, $w(0)=0$ in
(a), and $w(0)=0.01$ in (b).}
\label{experim1}
\end{figure}

In order to prove the soundness of the proposed theory we performed
some simulations of the Euler equations~\eqref{x-eq} and~\eqref{W-eq}
in the special case $\omega=1$, $\eta=1$, $\varpi=\exp(\vartheta t)$ and
$F=-\exp(\vartheta t) \overline V(t,x(t))$, where in particular
$\overline V(t,x(t))$ is taken to be a quadratic loss on the output neuron.
To understand the learning dynamic of the weights
we choose a constant supervision signal and various time-dependent
input signals $e(t)$.
Figure~\ref{experim1} shows the evolution of the weight of a single 
linear neuron $x(t)=w(t) e(t)$
with a target $y=3$ and a variable input $e(t)$. In
Fig.~\ref{experim1}--(a) $e(t)\to 3$ as $t\to\infty$, and indeed
$w(t)$ converges to $1$. In
Fig.~\ref{experim1}--(b) $e(t)\approx 3(1-t)$ and consistently $w(t)\approx
1/(1-t)$. Notice that in both cases the neuron constraint is always
exactly satisfied.
Remember that the initial conditions must be consistent with
Eq.~\eqref{initial-cond}; in this example in Fig.~\ref{experim1}--(a)
we have $w(0)=0$ that guaranteed $G_\tau=0$, while in the experiment
relative to Fig.~\ref{experim1}--(b) one can choose $\dot w(0)\ne 0$ as
the condition $G_\tau=0$ is ensured by $\dot e(0)=0$.

\begin{figure}
\hbox to\hsize{
\includegraphics[width=0.5\hsize]{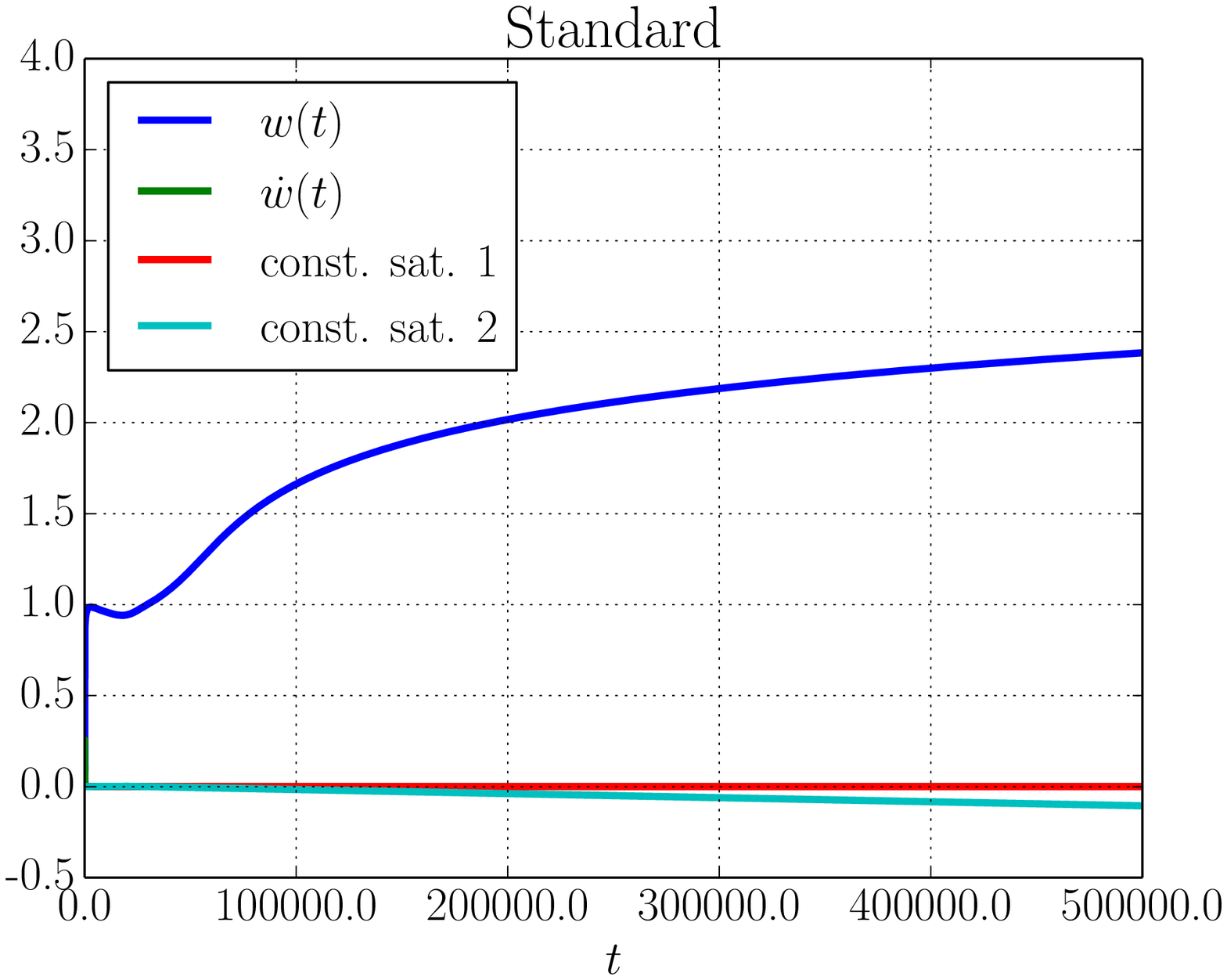}\hfill
\includegraphics[width=0.5\hsize]{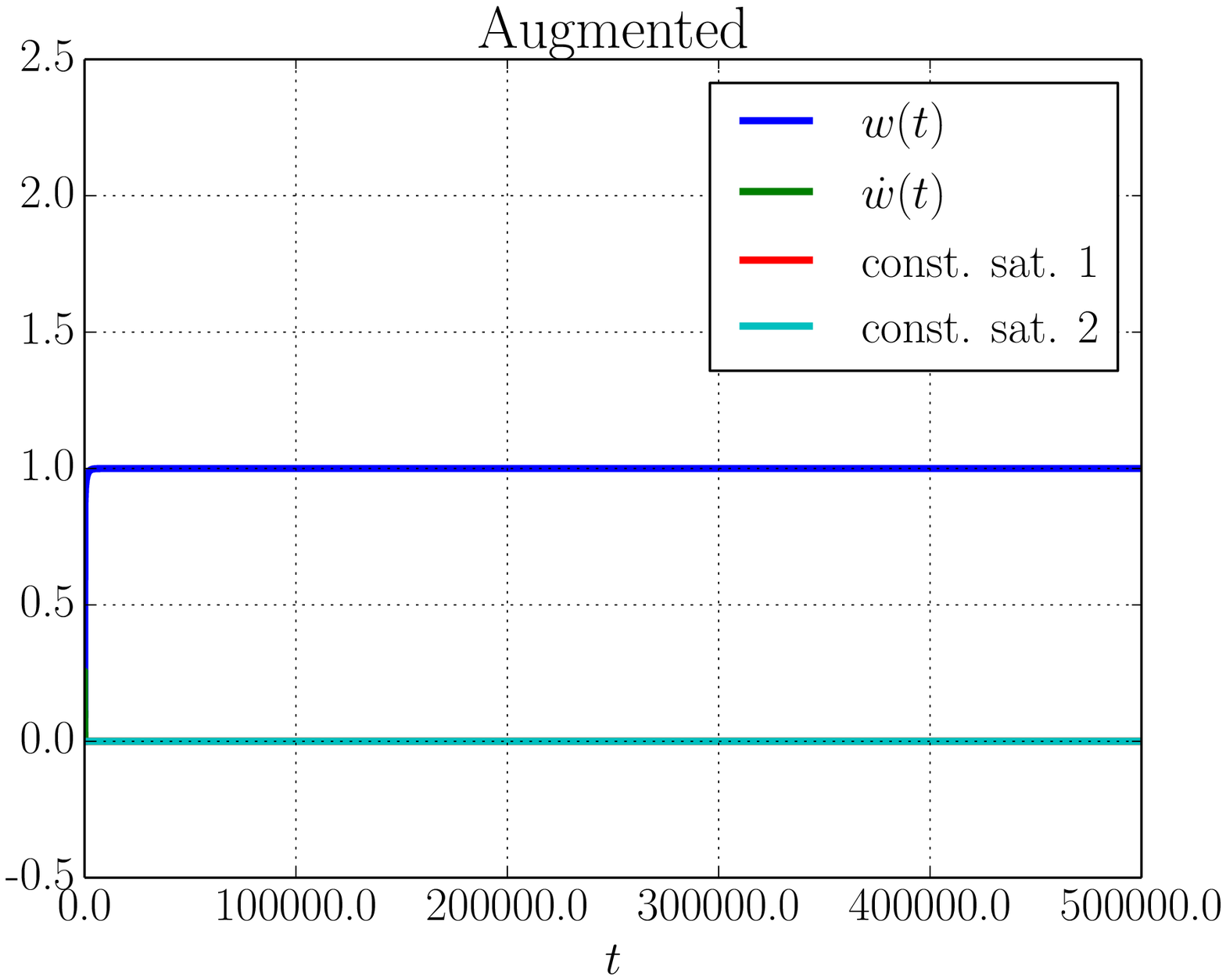}
}
\hbox to \hsize{\hfil(a)\qquad\quad\hfil\quad\qquad(b)\hfil}
\caption{Temporal evolution of the weight that connects two neurons:
$G^1=x^1-e$, $G^2=x^2-\sigma(w x^1)$ with $\sigma(x):=\Th(x)$. We took
$m_x=m_W=\vartheta=1$ and identically null initial conditions.
In (a) we considered $\overline V=(x^2-\Th(3))^2/2$, while in (b)
we chose $\overline V=(x^2-\Th(3))^2/2+(x^2-\sigma(w x^1))^2/2$. 
}
\label{experim2}
\end{figure}

In Fig.~\ref{experim2} instead  we tested the robustness of the method
with respect to numerical errors by running the simulation for a longer
period of time. The model here consists of  two neurons NN with nonlinear
activation function. We observed that due to numerical errors the system
can fail to converge to the correct solution $w=1$
(Fig.~\ref{experim2}--(a)). This can be understood as soon as we realize that,
following the ideas of Section~\ref{dirichlet-sec}, EL-equations
implements only the satisfaction of the second derivative of the
constraints, therefore errors on the trajectories can shift the dynamic
of the system on another constraint that differs from the correct one by
a linear function of time.
Hence, we found that such behaviour can be effectively
corrected (see Fig.~\ref{experim2}--(b))
by adding to the potential  a quadratic loss
on the constraint itself.



%% file: conclusions.tex
This paper proposes a novel formulation of learning 
by differential equations instead of by the dominating approach 
of using finite-dimensional optimization. 
This can be traced back many contributions that early
appeared at the of the eighties (see e.g.~\cite{Pineda87} and~\cite{Pearlmutter89b}), as well as from the
recent trend of emphasizing
continuous-based computational models of learning (see e.g.~\cite{conf/nips/ChenRBD18}\footnote{Best student paper awards at NeurIPS 2018.}.
The distinctive view proposed in this paper consists of 
the close parallel with mechanics, that arises from 
the general principle of formulating  variational laws of nature. 
The STLP computational scheme  possesses the distinctive 
feature of being local in both space and time. 
Moreover, the gained space locality property goes 
beyond the classic local neural communication required for computing
the gradient. Unlike BP,  there is no need to synchronize the 
forward and backward step that return the factors of the gradient,
since they are locally available.
The theory nicely addresses classic arguments 
on BP biologically plausibility~\cite{Grossberg87b}, 
and opens the doors to an
in-depth reformulation of learning algorithms for both
feedforward and recurrent neural networks.

%% file: neurips_2019-ccal.bbl
\begin{thebibliography}{10}

\bibitem{Rumelhart86a}
D.E. Rumelhart, J.L. McClelland, and the PDP Research~Group.
\newblock {\em Parallel Distributed Processing: Explorations in the
  Microstructure of Cognition}, volume~1.
\newblock MIT Press, Cambridge, 1986.

\bibitem{BottouCN18}
L{\'{e}}on Bottou, Frank~E. Curtis, and Jorge Nocedal.
\newblock Optimization methods for large-scale machine learning.
\newblock {\em {SIAM} Review}, 60(2):223--311, 2018.

\bibitem{Pineda87}
F.J. Pineda.
\newblock Generalization of back-propagation to recurrent neural networks.
\newblock {\em Physical Review Letters}, 59:2229--2232, 1987.

\bibitem{Pearlmutter89b}
B.A. Pearlmutter.
\newblock Learning state space trajectories in recurrent neural networks.
\newblock {\em Neural Computation}, 1:263--269, 1989.

\bibitem{conf/nips/ChenRBD18}
Tian~Qi Chen, Yulia Rubanova, Jesse Bettencourt, and David~K. Duvenaud.
\newblock Neural ordinary differential equations.
\newblock In Samy Bengio, Hanna~M. Wallach, Hugo Larochelle, Kristen Grauman,
  Nicol\`o Cesa-Bianchi, and Roman Garnett, editors, {\em NeurIPS}, pages
  6572--6583, 2018.

\bibitem{DBLP:journals/corr/BengioLBL15}
Yoshua Bengio, Dong{-}Hyun Lee, J{\"{o}}rg Bornschein, and Zhouhan Lin.
\newblock Towards biologically plausible deep learning.
\newblock {\em CoRR}, abs/1502.04156, 2015.

\bibitem{Williams89b}
R.J. Williams and D.~Zipser.
\newblock A learning algorithm for continually running fully recurrent neural
  networks.
\newblock {\em Neural Computation}, 1:270--280, 1989.

\bibitem{giaquita-hildebrandt}
Mariano Giaquinta and Stefan Hildebrandt.
\newblock Calculus of variations, vol. i. number 310 in a series of
  comprehensive studies in mathematics, 1996.

\bibitem{courant-hilbert}
Richard Courant and David Hilbert.
\newblock {\em Methods of Mathematical Physics, Vol. I.}
\newblock John Wiley \& Sons, 1991.

\bibitem{FraDah11}
E.~Frazzoli and M.~Dahleh.
\newblock Lectures on dynamic systems and control.
\newblock MIT course 6.241J / 16.338J.

\bibitem{Grossberg87b}
S.~Grossberg.
\newblock Competitive learning: From interactive activation to adaptive
  resonance.
\newblock {\em Cognitive Science}, 11:23--63, 1987.

\end{thebibliography}
